\documentclass[oribibl]{llncs}

\usepackage{makeidx}  % allows for indexgeneration
\usepackage[usenames,dvipsnames,svgnames,table]{xcolor}
\usepackage{mdframed}
\usepackage{amssymb}
\usepackage{amsmath}
\usepackage{proof}  
\usepackage{bussproofs}
\usepackage{verbatim}
\usepackage{url}
\usepackage{fancybox}

\usepackage[amsmath,thmmarks]{ntheorem}
\usepackage{relsize}
\usepackage{mathtools}    
\DeclarePairedDelimiterX\setc[2]{\{}{\}}{\,#1 \;\delimsize\vert\; #2\,}

\newcommand{\IGNORE}[1]{}

\newcommand{\lsort}[1]{%
  \ensuremath{\mbox{\textsf{#1}}}}
\newcommand{\defsort}[2]{%
  \newcommand{#1}{\lsort{#2}}}
\defsort{\Action}{Action}
\defsort{\Time}{Time}
\defsort{\Self}{Self}

\defsort{\Agent}{Agent}
\defsort{\Entrant}{Entrant}
\defsort{\ActionType}{ActionType}
\defsort{\Moment}{Moment}
\defsort{\Boolean}{Formula}
\defsort{\PayOut}{PayOut}
\defsort{\Fluent}{Fluent}
\defsort{\Event}{Event}
\defsort{\Object}{Object}
\defsort{\RealTerm}{RealTerm}

\defsort{\Numeric}{Numeric}
\defsort{\Number}{Number}
\defsort{\Trolley}{Trolley}
\defsort{\Track}{Track}
\defsort{\Moveable}{Moveable}
\defsort{\Situation}{Situation}

\newcommand{\lsymbol}[1]{%
  \ensuremath{\mathit{#1}}}
\newcommand{\defsymbol}[2]{%
  \newcommand{#1}{\lsymbol{#2}}}
\defsymbol{\action}{action}
\defsymbol{\initially}{initially}
\defsymbol{\holds}{Holds}
\defsymbol{\happens}{happens}
\defsymbol{\clipped}{clipped}
\defsymbol{\initiates}{initiates}
\defsymbol{\terminates}{terminates}
\defsymbol{\prior}{prior}
\defsymbol{\interval}{interval}

\defsymbol{\does}{does}
\defsymbol{\plans}{plans}
\defsymbol{\act}{act}
\defsymbol{\react}{react}
\defsymbol{\payTot}{pay_{tot}}
\defsymbol{\fight}{fight}
\defsymbol{\coop}{coop}
\defsymbol{\enter}{enter}
\defsymbol{\stayout}{stayout}
\defsymbol{\learns}{learns}
\defsymbol{\payoff}{payoff}
\defsymbol{\position}{position}
\defsymbol{\dead}{dead}
\defsymbol{\damaged}{damaged}

\defsymbol{\onrails}{onrails}
\defsymbol{\switch}{switch}
\defsymbol{\drop}{drop}
\defsymbol{\ins}{in}
\defsymbol{\actionsit}{Action}
\defsymbol{\sick}{sick}
\defsymbol{\visit}{visit}

\newcommand{\lconstant}[1]{%
  \ensuremath{\mbox{\textsf{#1}}}}
\newcommand{\defconstant}[2]{%
  \newcommand{#1}{\lconstant{#2}}}
\defconstant{\Enter}{Enter}
\defconstant{\StayOut}{StayOut}
\defconstant{\Fight}{Fight}
\defconstant{\Acquiesce}{Acquiesce}
\defconstant{\cs}{cs }

\defconstant{\john}{John}
\defconstant{\doctor}{doctor}

\newcommand{\lmodality}[1]{%
  \ensuremath{\mathbf{#1}}}
\newcommand{\defmodality}[2]{%
  \newcommand{#1}{\lmodality{#2}}}
\defmodality{\common}{C}
\defmodality{\knows}{K}
\defmodality{\believes}{B}
\defmodality{\perceives}{P}
\defmodality{\mental}{M}

\defmodality{\desires}{D}
\defmodality{\intends}{I}
\defmodality{\says}{S}
\defmodality{\ought}{O}

\mathchardef\mhyphen="2D

\newcommand{\sep}{\ \lvert \ }

\usepackage{booktabs}
\usepackage{mdframed}
\usepackage{paralist}

\usepackage{color}

\usepackage[usenames,dvipsnames,svgnames,table]{xcolor}

\newcommand{\mCp}{\ensuremath{{\mathcal{{\mu C}}}}}

\begin{document}

\title{Toward Formalizing Teleportation of Pedagogical Artificial Agents}

\author{John Angel \and Naveen Sundar Govindarajulu \and Selmer Bringsjord}

\authorrunning{John Angel et al.}

\institute{Rensselaer AI \textit{\&} Reasoning (RAIR) Lab\\
           Department of Computer Science \& Department of Cognitive Science\\
           Rensselaer Polytechnic Institute (RPI) $\bullet$ Troy NY 12180 USA}

\date{\today}

\maketitle

\begin{comment}
\begin{abstract}
\noindent
%
Our paradigm for the use of artificial agents to teach requires among
other things that they persist through time in their interaction with
human students, in such a way that they ``teleport'' or ``migrate''
from an embodiment at one time $t$ to a different embodiment at later
time $t'$.  In this short paper, we report on initial steps toward the
formalization of such teleportation, in order to enable an overseeing
AI system to establish, mechanically, and verifiably, that the human
students in question will likely believe that the very same artificial
agent has persisted across such times despite the different
embodiments.
\end{abstract}
\end{comment}

\vspace{-0.07in}

\section{Introduction}
\label{sect:intro}

Our paradigm for the use of artificial agents to teach requires among
other things that they persist through time in their interaction with
human students, in such a way that they ``teleport'' or ``migrate''
from an embodiment at one time $t$ to a different embodiment at later
time $t'$.  In this short paper, we report on initial steps toward the
formalization of such teleportation, in order to enable an overseeing
AI system to establish, mechanically, and verifiably, that the human
students in question will likely believe that the very same artificial
agent has persisted across such times despite the different
embodiments.

The plan for the sequel is straightforward, and as follows.  After
encapsulating our paradigm for the deployment of artificial agents in
service of learning, and taking note of the fact that the
``teleportation''/``migration'' problem has hitherto been treated only
informally, we then convey the kernel of our approach to formalizing
agent teleportation between different embodiments, formalize this
kernel to a degree, in order to produce an initial simulation, and
then wrap up with some final remarks.

\vspace{-0.07in}

\section{Our Paradigm \& Teleportation}
\label{sect:tippae_summary}

A crucial part of our novel paradigm for artificial agents that teach
is the engineering
%% , and exploring the real-world, ethical ramifications
of a class of AIs, crucially powered by cognitive logics, able to
persist through days and weeks in their interaction with the humans
whose education is to be thereby enhanced.  The artificial agents in
our paradigm are able to seamlessly ``teleport'' between heterogenous
environments in which a human learner may find herself as time
unfolds; this capacity is intended to provide a continuous educational
experience to the human student, and offers the possibility of
human-machine friendship.

In short, our agents need to be ``teleportative.''  This means that
the agent should be able to be used in multiple hardware environments
by a user, such that the user has the impression of an continuous,
uninterrupted interaction with the very same agent.  
%% Furthermore, the user of a teleportative, or migrating, agent
%% should have the impression that the agent in one hardware
%% environment is the same agent that they have been interacting with
%% after the act of teleporting has occurred.
This helps to reinforce the possibility of a persistent, trusting
relationship between human and machine.

\vspace{-0.07in}

\section{Prior Accounts of Teleportation of Artificial Agents}
\label{sect:prior_art}

There is some excellent and interesting prior work on teleporting
artificial agents.  Some explore how the consistency of a migrating
agent's memory effects a user's perception of a continuous identity
\cite{do_I_remember_you}.  Others shed light on visual cues useful for
convincing users of an agent's teleportation \cite{Koay2009}.  In
addition, excellent progress has been made towards the designing of
migrating agents \cite{Hassani2014} and testing real-world
implementations of such agents \cite{Gomes2011}.  Unfortunately for
our purposes, the prior art is informal.  Our goal is to capture
teleportation formally, and on the strength of that formalization to
enable an overseeing AI system to prove, or minimally justify
rigorously, that the teleportation in question is indeed believable.
%% (based on the memory-based conception of personal identity through
%% time, which is cited shortly).
%
%% JOHN: Expand this citation to include the additional three papers,
%% which you will need to give BibTex entries for in your bib file.
%% And then hopefully you can say a few words about the four papers
%% being good, but again informal.  Thx!  //Selmer

\section{The Kernel of the Formalization}
\label{sect:kernel}

In the longstanding quasi-technical literature on \textit{personal
  identity} in philosophy there is a strong tradition of trying to
work out a rigorous account of when person $p_1$ at $t_1$ (=
$p_{t_1}$) is identical with person $p_{t_2}$ on the basis of shared
memories between $p_{t_1}$ and $p_{t_2}$.
%% Memories aren't infallible; they are \textbf{evidently true}.

The goal of our initial formalization is to build a system that can
find a proof for when it believes that a student believes two embodied
agents are the same $p_{t_1} \equiv p_{t_2}$.  The system can conclude
that the student believes two embodiments to be the same if the system
can find a proof that it believes that the student believes that the
two embodiments have a belief $\beta$ at specific times that cannot be
believed by more than one agent.  If the system fails to find such a
proof or argument, then the system can take corrective actions to make
it more explicit to the human that the embodiments are the same.  Note
that formalization requires the system to understand beliefs of agents
which might themselves be about beliefs of other agents (and so on).

%% (This is a technical category that you aren't yet
%% familiar with, John.)  

%%%%%%%%%%%%%%%%%%%%%%%%%%%%%%%%%%%%%%%%%%%%%%%%%%%%%%%%%%%%%%%%%%%%%%
\begin{comment}

Now, that would mean in an uncertainty-ized cognitive calculus that

$$\mathbf{B}^{12}(p_{t_1} \phi[p_{t_1}])$$

\noindent
%
i.e.\ that $p_{t_1}$ believes at level 12 that $\phi$, which involves
$p_{t_1}$, holds; and it would also mean that

$$\mathbf{B}^{12}(p_{t_2} \phi[p_{t_2}]),$$

\noindent
%
where otherwise the two formulae $\phi[p_{t_1}]$ and $\phi[p_{t_1}]$
are syntactically identical.

Let's suppose that this can be worked out in a simple scenario within
a cognitive calculus.  Naveen :)?  Okay, now, my idea is this: Suppose
we have a human agent $h$ in the picture, and that $p_{t_1}$ and
$p_{t_2}$ here aren't persons, but rather artificial agents (AIs).  We
can say (or conjecture or propose; and I don't know if this is in line
with the papers John sent; would be great if it is) that $h$ believes
(at some level above more-probable-than not) that $a_{t_1}$ and
$a_{t_2}$ (each with different embodiments) are identical if and only
if this pair has shared memories as above.  In this case, $h$ would
have perceived information at $t_1$1 (which leads to a belief on $h$'s
part that $a_{t_1}$ has memories) and perceived information at $t_2$
(which leads to a belief on $h$'s part that $a_{t_2}$ has the same
memories).

\end{comment}
%%%%%%%%%%%%%%%%%%%%%%%%%%%%%%%%%%%%%%%%%%%%%%%%%%%%%%%%%%%%%%%%%%%%%%

\vspace{-0.07in}

\section{Initial Formalization and Simulation}
\label{sect:initial_formalization_sim}

The requirement that the system understand the student's beliefs about
other embodied agents' beliefs implies that we need to have a
sufficiently expressive system 
%% to build our formalization on. 
\textbf{BDI logics} (belief/desire/intentions) have a long tradition
of being used to model such agents \cite{wooldridge.mas}.  For our
formalization, we use a system that is a proper superset\footnote{With
  respect to modal operators and inference schemata.} of such logics.

%% \subsection{Calculus} 
%% \label{subsect:cog_calc_shadowprover}

 We specifically use the formal system $\mCp$ in
 \cite{govindarajulu2017strength}.  The system is a modal extension of
 first-order logic, and has the following modal operators:
 $\believes$, for belief, and $\perceives$ for perception.  The syntax
 and inference schemata of the system are shown below.  $\phi$ is a
 meta-variable for formulae, and $A(\ldots)$ is any first-order atomic
 formula.  Assume that we have at hand a first-order alphabet
 augmented with a finite set of constant symbols for agents $\{a_1,
 \ldots, a_n\}$ and a countably infinite set of constant symbols for
 times $\{t_0, t_1, \ldots \}$. (Sometimes we use $a$ for $a_i$ and
 $t$ for $t_i$ below.) $\{x, y, z,\ldots \}$ are first-order
 variables. The grammar is follows:

\begin{scriptsize}
 \begin{equation*}
 \begin{aligned}
     \mathit{\phi}&::= \left\{ 
    \begin{aligned}
     &A(\ldots) \sep  \neg \phi \sep \phi \land \psi \sep \phi \lor
     \psi \sep \phi \rightarrow \psi \sep \phi \leftrightarrow \psi
     \sep \forall x. \phi  \perceives (a,t,\phi) \sep \believes (a,t,\phi)  %\sep \desires(a,t,\phi) \sep \intends(a,t,\phi) \sep \ought(a,t,\phi, \psi) 
      \end{aligned}\right.
  \end{aligned}
\end{equation*}
 \end{scriptsize}
$\believes(a, t, \phi)$ stands for agent $a$ at time $t$ believing
$\phi$ and $\perceives(a, t, \phi)$ stands for agent $a$ at time $t$
perceiving $\phi$. The base inference schemata are given here:

\begin{scriptsize}
  \begin{equation*}
\begin{aligned}
 &\hspace{35pt}\mbox{All propositional natural-deduction inference schemata.} [R_{\mathbf{ND}_0}]\\
&\mbox{Natural-deduction quantifier inference schemata treating modal formulae
  opaquely} [R_{\mathbf{ND}_1}]\\
&\hspace{30pt}  \infer[{[R_{\perceives}]}]{\believes(a,t_2,\phi)}{\perceives(a,t_1,\phi_1),
    \ \ \Gamma \vdash t_1 < t_2} \\
  &\infer[{[R_{\believes}]}]{\believes(a,t,\phi)}{\believes (a,t_1,\phi_1),
    \ldots, \believes (a,t_m,\phi_m), \{\phi_1, \ldots, \phi_m\}\vdash
    \phi,  \Gamma\vdash t_i < t} 
\end{aligned}
\end{equation*}
 \end{scriptsize}

 Now assume that there is a background set of axioms $\Gamma$ we are
 working with.  We have available the basic theory of arithmetic (so
 we can assert statements of the form $x < y$).  $R_{\mathbf{P}}$
 tells us how perceptions get translated into beliefs.
 $R_{\mathbf{B}}$ is an inference schema that lets us model idealized
 agents that have their beliefs closed under the \mCp\ proof theory.
 While normal humans are not deductively closed, this lets us model
 more closely how deliberate agents such as organizations and more
 strategic actors reason.  Reasoning is performed through a novel
 first-order modal logic theorem prover, \textsf{ShadowProver}, which
 uses a technique called \textbf{shadowing} to achieve speed without
 sacrificing consistency in the system
 \cite{nsg_sb_dde_2017}.\footnote{The prover is available in both Java
   and Common Lisp and can be obtained at:
   \url{https://github.com/naveensundarg/prover}. The underlying
   first-order prover is SNARK, available at:
   \url{http://www.ai.sri.com/~stickel/snark.html}.}

%% \subsection{Simulation} 

%%%%%%%%%%%%%%%%%%%%%%%%%%%%%%%%%%%%%%%%%%%%%%%%%%%%%%%%%%%%%%%%%
\begin{figure}

\centering
\caption{Simulation}  \medskip
  \label{fig:diagram}
\shadowbox{\includegraphics[width=0.45\linewidth]{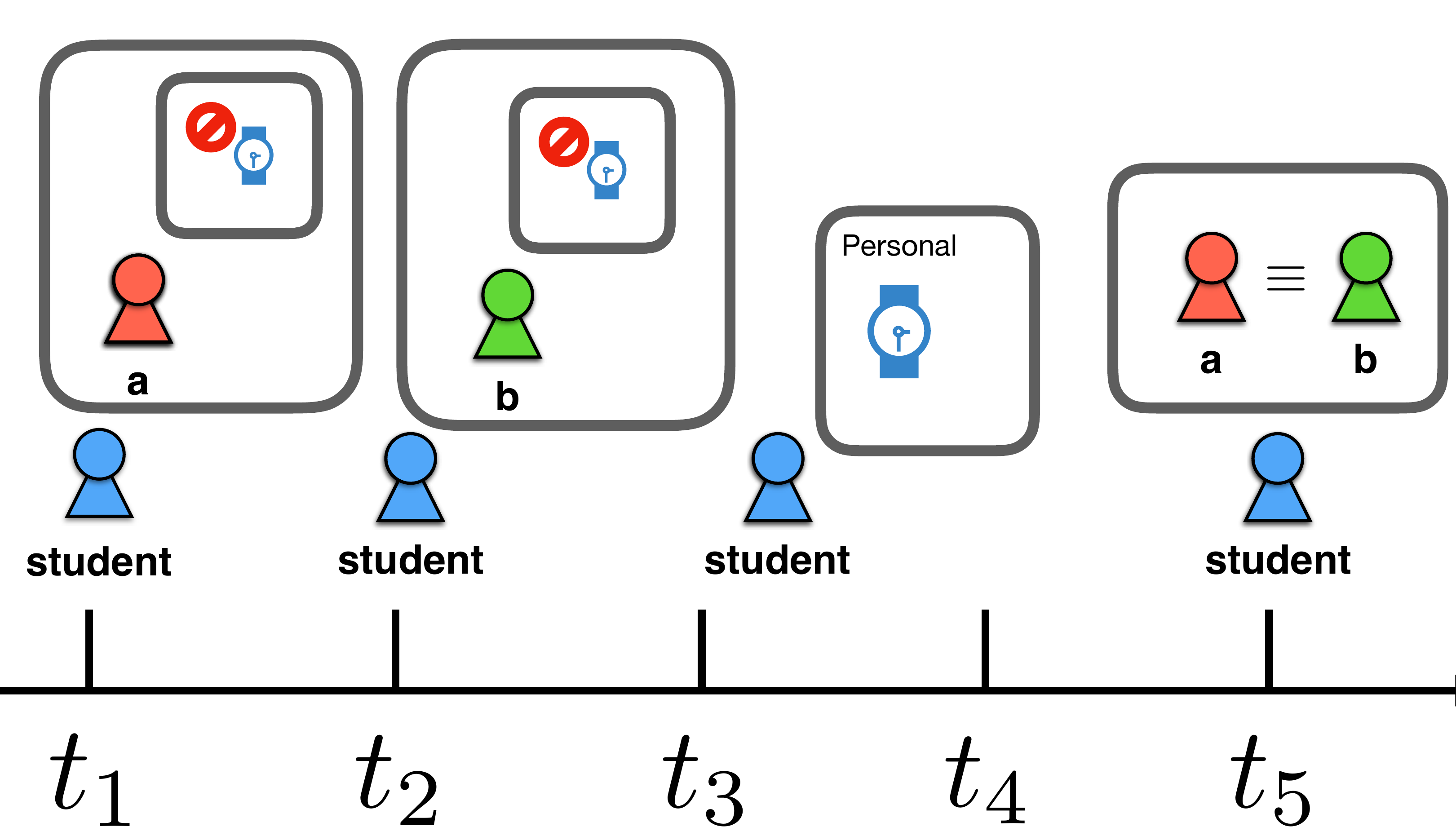}}
\end{figure}
%%%%%%%%%%%%%%%%%%%%%%%%%%%%%%%%%%%%%%%%%%%%%%%%%%%%%%%%%%

\vspace{-0.20in} 
The simulation is set up as a reasoning problem from a set of given
assumptions to a goal (see Figure~\ref{fig:proof1}).  In the
formalization shown below, the system believes that the student
believes two embodiments to have the same identity \emph{if} the
embodiments at different times believe some personal object to have
the same property (see assumption $\mathsf{A4}$).  For instance,
assume that the student's watch is a personal object.  At time $t_1$,
we have $\mathsf{(embodiment \ a)}$ believing that the watch is
stopped, and at time $t_2$ we also have $\mathsf{(embodiment\ b)}$
believing the same.  From these assumptions, the system can derive
that the student believes that the embodiments are the same (see
Figure~\ref{fig:diagram} for an overview).

%%%%%%%%%%%%%%%%%%%%%%%%%%%%%%%%%%%%%%%%%%%%%%%%%%%%%%%%%%%%%%%%%
\begin{figure}
\centering
 \caption{Simulation: \textsf{ShadowProver} goes through the below in $\approx 2.8$ seconds.}  \medskip
  \label{fig:proof1}
\shadowbox{\includegraphics[width=0.80\linewidth]{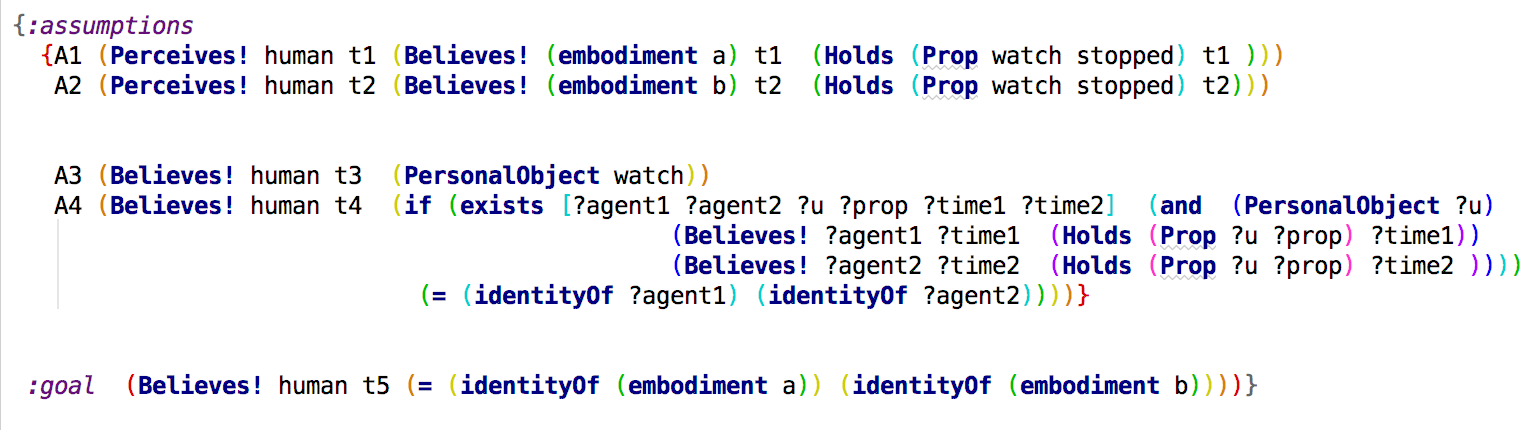}}
 \end{figure}
%%%%%%%%%%%%%%%%%%%%%%%%%%%%%%%%%%%%%%%%%%%%%%%%%%%%%%%%%%

\section{Concluding Remarks; Next Steps}
\label{sect:final_remarks}

We readily admit to having only taken initial steps \emph{toward} the
formalization of teleportation for artificial agents.  The simulation
we have presented does seem to indicate to us
%% , in no small part
%% because of the robustness of the computational logics and implemented
%% systems upon which it's based, 
that things are scalable --- but of course only time and
experimentation will tell.  Finally, it's important to note that we
haven't herein sought to address the educational \emph{efficacy} of
our approach, nor the specific learning value of persistent teaching
agents across embodiments.
%%  Yet, without a formal account of teleportation, and corresponding,
%% systematic confidence that teleportation of the right structure
%% preserves identity through time, educational efficacy would be
%% precluded in any case.

\vspace{-0.15in}

\begin{small}
  \bibliography{main72,naveen,john}

\begin{thebibliography}{1}
\providecommand{\url}[1]{\texttt{#1}}
\providecommand{\urlprefix}{URL }

\bibitem{do_I_remember_you}
Aylett, R., Kriegel, M., Wallace, I., Segura, E.M., Mecurio, J., Nylander, S.,
  Vargas, P.: {Do I Remember You? Memory and Identity in Multiple Embodiments}.
  In: Proceedings of The 22nd IEEE International Symposium on Robot and Human
  Interactive Communication (RO-MAN 2013). {IEEE} (2013), {DOI:
  10.1109/ROMAN.2013.6628435. The conference was located in Gyeongju, South
  Korea.}

\bibitem{Gomes2011}
Gomes, P.F., Segura, E.M., Cramer, H., Paiva, T., Paiva, A., Holmquist, L.E.:
  {ViPleo and PhyPleo: Artificial Pet with Two Embodiments}. Proceedings of the
  8th International Conference on Advances in Computer Entertainment Technology
  pp. 3:1--3:8 (2011), \url{http://doi.acm.org/10.1145/2071423.2071427}

\bibitem{nsg_sb_dde_2017}
Govindarajulu, N.S., Bringsjord, S.: {On Automating the Doctrine of Double
  Effect}. In: Sierra, C. (ed.) {Proceedings of the Twenty-Sixth International
  Joint Conference on Artificial Intelligence, {IJCAI-17}}. pp. 4722--4730.
  Melbourne, Australia (2017), \url{https://doi.org/10.24963/ijcai.2017/658},
  preprint available at this url: \url{https://arxiv.org/abs/1703.08922}

\bibitem{govindarajulu2017strength}
Govindarajulu, N.S., Bringsjord, S.: {Strength Factors: An Uncertainty System
  for a Quantified Modal Logic} (2017), \url{https://arxiv.org/abs/1705.10726},
  presented at Workshop on Logical Foundations for Uncertainty and Machine
  Learning at IJCAI 2017, Melbourne, Australia

\bibitem{Hassani2014}
Hassani, K., Lee, W.S.: {On designing migrating agents}. SIGGRAPH Asia 2014
  Autonomous Virtual Humans and Social Robot for Telepresence on - SIGGRAPH
  ASIA '14 pp. 1--10 (2014),
  \url{http://dl.acm.org/citation.cfm?doid=2668956.2668963}

\bibitem{Koay2009}
Koay, K.L., Syrdal, D.S., Walters, M.L., Dautenhahn, K.: {A User study on
  visualization of agent migration between two companion robots.} HCII '09:
  Proceedings of the 13th International Conference on Human-Computer
  Interaction  (2009), \url{http://uhra.herts.ac.uk/handle/2299/3977}

\bibitem{wooldridge.mas}
Wooldridge, M.: An Introduction to Multi Agent Systems. MIT Press, Cambridge MA
  (2002)

\end{thebibliography}
  \bibliographystyle{splncs03}
\end{small}

\end{document}